%% file: main.tex
\documentclass[runningheads]{llncs}

\usepackage{eccv}
\usepackage{eccvabbrv}
\usepackage{graphicx}
\usepackage{booktabs}
\usepackage[accsupp]{axessibility}
\usepackage{placeins}
\usepackage{multirow}
\usepackage[table]{xcolor}
\usepackage{float}
\usepackage{xcolor}

\usepackage{hyperref}

\usepackage{orcidlink}

\begin{document}

\title{TAG: Target-Agnostic Guidance for Stable Object-Centric Inference in Vision-Language-Action Models} 
\titlerunning{Target-Agnostic Guidance for VLA}

\author{
  Jiaying Zhou\inst{1,*}\orcidlink{0009-0002-0563-3483} \and
  Zhihao Zhan\inst{1,*}\orcidlink{0009-0001-4757-260X} \and
  Ruifeng Zhai\inst{1} \and
  Qinhan Lyu\inst{1}\orcidlink{0009-0008-7188-6802} \and
  Hao Liu\inst{1}\orcidlink{0009-0003-5541-3558} \and \\
  Keze Wang\inst{1,2,3}\orcidlink{0000-0002-7817-8306} \and
  Liang Lin\inst{1,2,3}\orcidlink{0000-0003-2248-3755}  \and
  Guangrun Wang \inst{1,2,3,\dagger}\orcidlink{0000-0001-7760-1339}
}

\authorrunning{Zhou et al.}

\institute{Sun Yat-sen University \and
Guangdong Key Lab of Big Data Analysis \& Processing \and
X-Era AI Lab \\
\email{
\{\href{mailto:zhoujy97@mail2.sysu.edu.cn}{zhoujy97},\href{mailto:zhanzhh6@mail2.sysu.edu.cn}{zhanzhh6}\}@mail2.sysu.edu.cn,
\href{mailto:wanggrun@gmail.com}{wanggrun@gmail.com}}
}

\begingroup
   \renewcommand\thefootnote{\fnsymbol{footnote}} 
   \footnotetext{$*$ {Equal contribution}.}
   \footnotetext{$\dagger$ {Corresponding author}.}
\endgroup

\maketitle

\input{sec/0_abstract}
\input{sec/1_intro}

\input{sec/2_relate}

\input{sec/3_method}
\input{sec/4_exp}
\input{sec/5_conclusion}

\bibliographystyle{plain}
\bibliography{main}

\input{sec/X_suppl}

\end{document}

%% file: sec/0_abstract.tex
\begin{abstract}

Vision--Language--Action (VLA) policies have shown strong progress in mapping language instructions and visual observations to robotic actions, yet their reliability degrades in cluttered scenes with distractors. 
By analyzing failure cases, we find that many errors do not arise from infeasible motions, but from instance-level grounding failures: the policy often produces a plausible grasp trajectory that lands slightly off-target or even on the wrong object instance. 
To address this issue, we propose \textbf{TAG} (Target-Agnostic Guidance), a simple inference-time guidance mechanism that explicitly reduces distractor- and appearance-induced bias in VLA policies. 
Inspired by classifier-free guidance (CFG), TAG contrasts policy predictions under the original observation and an object-erased observation, and uses their difference as a residual steering signal that strengthens the influence of object evidence in the decision process. 
TAG does not require modifying the policy architecture and can be integrated with existing VLA policies with minimal training and inference changes.
We evaluate TAG on standard manipulation benchmarks, including LIBERO, LIBERO-Plus, and VLABench, where it consistently improves robustness under clutter and reduces near-miss and wrong-object executions.

\keywords{Vision-Language-Action Models \and Robotics \and Target-Agnostic Guidance}
\end{abstract}

%% file: sec/1_intro.tex
\section{Introduction}

Vision–Language–Action (VLA) models have recently shown promising results in mapping natural-language instructions and visual observations to robotic actions. 
However, failures follow a consistent pattern: rather than producing infeasible motions, policies often execute a plausible grasp yet land slightly off-target—sometimes even on the wrong object instance—in cluttered scenes with nearby distractors. This motivates a closer look at how clutter interferes with instance-level grounding.

From a visual standpoint, the observation contains multiple sources of evidence: (1) object-centric cues that specify where to act, (2) contextual cues from other manipulable items that define relative spatial layout, and (3) background appearance statistics such as textures, reflections, and lighting style. 
In principle, instance-level grounding should be driven primarily by foreground object evidence.
However, in cluttered scenes, decisions can be influenced by competing candidates and by background appearance correlations learned from data. 
Importantly, while contextual objects provide useful spatial reference, they also introduce competition in the candidate set. 
The challenge, therefore, is not to eliminate context altogether, but to suppress distractor- and appearance-induced bias while preserving the structural cues necessary for robust target association.

In most existing VLA policies, such foreground--background interactions are not explicitly modeled or regulated; instead, the policy is expected to learn from data how to balance object evidence, contextual structure, and background appearance within a unified representation (e.g., attention-based multimodal fusion in generalist VLA policies and diffusion/flow-style action generators)~\cite{kim2024openvla,mees2024octo,hou2025dita,chi2023diffusion,liu2024rdt,black2024pi_0}. 
While this implicit learning paradigm can be effective in nominal settings, recent robustness analyses report substantial performance degradation under controlled perturbations and visually challenging conditions (e.g., changes in lighting and background texture)~\cite{fei2025liberoplus,zhou2025libero,chen2026radar}. 

To stabilize instance grounding under clutter, we introduce TAG (Target-Agnostic Guidance), an inference-time guidance mechanism that explicitly reduces background- and distractor-induced bias in VLA policies.
TAG contrasts the policy’s predictions under the original observation and a counterfactual target-agnostic visual baseline that removes target evidence while preserving static scene structure (e.g., layout and surrounding objects), and uses their difference as a residual correction to make decisions more strongly driven by object evidence.
Inspired by classifier-free guidance (CFG), TAG amplifies this residual with a guidance scale, providing a simple knob to trade off robustness and fidelity without changing the policy architecture.

To make the counterfactual branch well-defined, we stochastically substitute a small fraction of target-erased observations during training, while using a temporally stable target-agnostic baseline at inference for a consistent two-branch contrast. 
This deliberate train--test design preserves physically meaningful context during learning and yields a cleaner nuisance estimate for guidance at deployment.
We evaluate TAG on standard manipulation benchmarks and show improved robustness to clutter and distractors, with substantial reductions in near-miss offsets and wrong-object executions.

\textbf{In summary, our contributions are threefold:}
\begin{enumerate}
    \item We identify instance-level target grounding under visual distraction as a dominant robustness bottleneck in VLA policies, characterized by systematic near-miss and wrong-object executions rather than motion infeasibility.
    
    \item We propose TAG (Target-Agnostic Guidance), a lightweight inference-time visual guidance mechanism inspired by CFG that explicitly suppresses distractor- and appearance-induced bias via residual contrast between target-present and counterfactual observations.
    
    \item We validate TAG on LIBERO~\cite{liu2023libero},  LIBERO-Plus~\cite{fei2025liberoplus}, VLABench~\cite{zhang2024vlabench}, demonstrating consistent improvements in success rate and visual robustness across diverse tasks and VLA architectures.
\end{enumerate}

%% file: sec/2_relate.tex
\section{Related Work}  \label{sec:relate}

\subsection{Vision-Language-Action Models}
Vision-Language-Action (VLA) architectures have rapidly evolved from pioneering large-scale imitators, such as RT-1~\cite{brohan2022rt} and RT-2~\cite{zitkovich2023rt}, to efficient open-source autoregressive frameworks exemplified by OpenVLA~\cite{kim2024openvla} and~\cite{song2026learning,song2025physical}. This autoregressive foundation has been further enriched by introducing explicit 3D spatial cues (SpatialVLA~\cite{qu2025spatialvla}) and optimizing continuous action fine-tuning paradigms (OpenVLA-OFT~\cite{kim2025fine}). To overcome the compounding errors and discretization bottlenecks inherent in autoregressive decoding, the generative paradigm has progressively shifted toward continuous action spaces. This transition spans from the seminal Diffusion Policy~\cite{chi2023diffusion} and cross-modal fusion designs like RDT~\cite{liu2024rdt}, to state-of-the-art flow-matching foundation models (e.g., $\pi_0$~\cite{black2024pi_0}, $\pi_{0.5}$~\cite{intelligence2025pi05}, and~\cite{zhan2025E0,li2025vla}.) and dual-system architectures (GR00T N1~\cite{bjorck2025gr00t}). Concurrently, to address the planning challenges in long-horizon tasks, researchers have augmented policies with explicit reasoning capabilities, incorporating textual Chain-of-Thought (CoT-VLA~\cite{zhao2025cot}) and visual generative rollouts (GR-1~\cite{wu2023unleashing}, GR-2~\cite{cheang2024gr}) to guide physical execution. Despite these massive strides in architectural scaling and logical reasoning across both autoregressive and diffusion-based frameworks, accurately mapping high-dimensional visual observations to robust actions in cluttered environments remains a fundamental challenge. Current policies still exhibit high sensitivity to local visual perturbations, often suffering from attention deviation and execution failures in the presence of task-irrelevant distractors.

\subsection{Classifier-Free Guidance in VLA Models}
Classifier-Free Guidance (CFG)~\cite{ho2022classifierfreediffusionguidance} has demonstrated remarkable efficacy in enhancing sample fidelity and conditional control in large generative models.
In the context of Vision-Language-Action (VLA) models, existing applications primarily focus on the language modality, where instructions are randomly dropped during training to strengthen semantic alignment during inference~\cite{zhan2026stablelanguageguidancevisionlanguageaction, fang2026when}. While some prior works have explored visual observation manipulation (e.g., dynamic token pruning in ADP~\cite{pei2025actionawaredynamicpruningefficient}), these approaches are primarily designed to reduce computational redundancy rather than being formulated as probabilistic guidance signals. Explicitly leveraging the CFG framework to decouple static backgrounds from manipulation targets remains underexplored. Our work fills this gap by introducing Target-Agnostic Guidance (TAG), extending the CFG paradigm from linguistic nullification to spatial-visual disentanglement.

%% file: sec/3_method.tex
\section{Method}
\begin{figure}[htbp]
  \centering
  \includegraphics[width=1\textwidth]{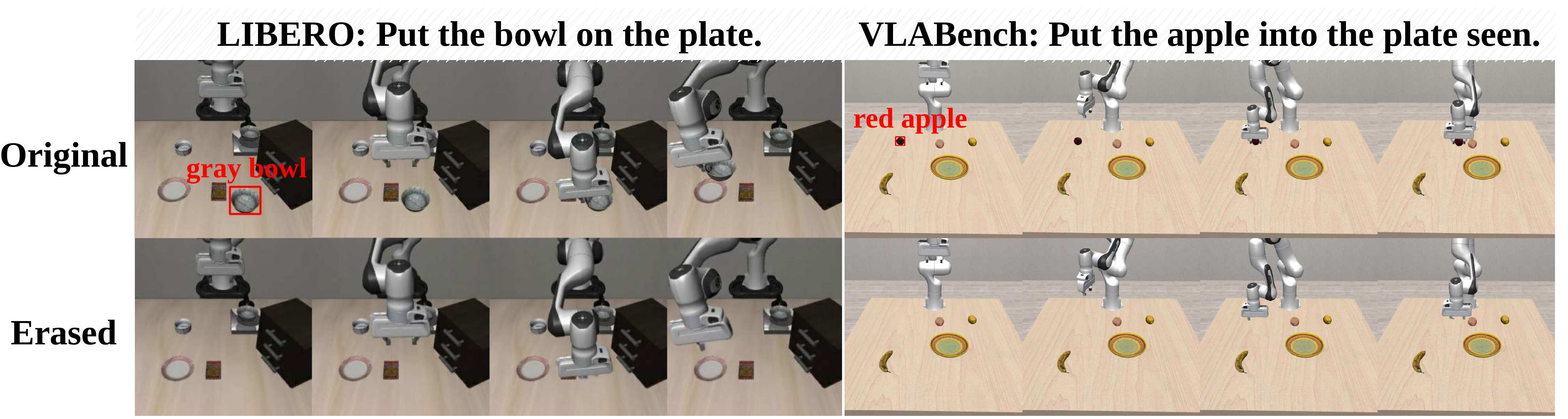}
  \caption{\textbf{Qualitative comparison of original and erased visual observations.} The top row displays original robotic manipulation trajectories from the Libero and VLABench datasets. The bottom row presents the corresponding sequences where the target objects (e.g., the bowl and apple) have been digitally erased. 
  }
  \label{fig:erased_img}
\end{figure}

\begin{figure}[t]
  \centering
  \includegraphics[width=1\textwidth]{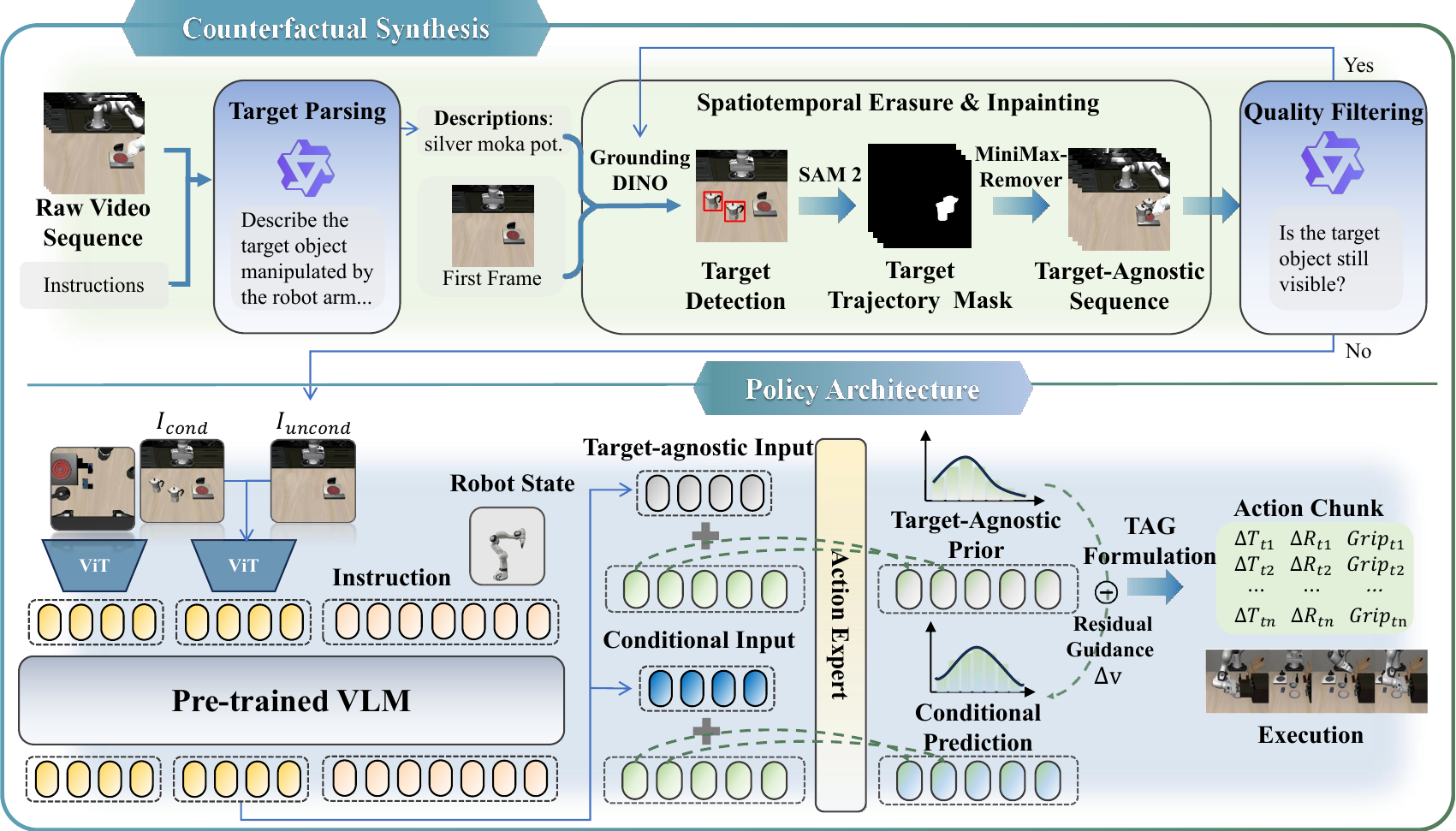}
  \caption{\textbf{Overview of the Target-Agnostic Guidance (TAG) framework. }  Our method consists of two core components. The top shows the Counterfactual Synthesis pipeline, which automatically erases the manipulated target from raw videos using target parsing, tracking, and inpainting to generate unconditional data. The bottom illustrates the Policy Architecture, where a dual-branch network processes both the original conditional input and the erased unconditional input. By extracting the residual guidance ($\Delta v$) between them, TAG effectively isolates background distractors, forcing the model to focus precisely on the target object and intended manipulation to generate accurate action sequences.
  }
  \label{fig:pipeline}
  
\vspace{-11pt}
\end{figure}

We propose \textbf{Target-Agnostic Guidance (TAG)}, an inference-time residual guidance mechanism to improve instance-level grounding under visual clutter.
TAG evaluates the same policy under a target-present observation $I_{\text{cond}}$ and a target-agnostic baseline $I_{\text{uncond}}$, and forms a guided prediction by CFG-style residual extrapolation (\cref{sec:tag_formulation}).
Unlike text conditioning, an ''unconditional'' visual input is not uniquely defined; thus, the key design dimension of TAG is how to construct $I_{\text{uncond}}$ to suppress target-specific nuisance cues while retaining the scene context needed for feasible control (\cref{sec:uncond_design}).

\subsection{Preliminaries}
Classifier-Free Guidance (CFG)~\cite{ho2022classifierfreediffusionguidance} is a widely used mechanism to improve controllability in conditional generative modeling by trading off condition adherence and sample diversity at inference time.

In a standard text-conditional CFG for diffusion models, the model evaluates both a conditional predictor $\epsilon_\theta(x_t, c)$ and an unconditional predictor $\epsilon_\theta(x_t, \emptyset)$, and performs residual extrapolation with a guidance scale $w$:
\begin{equation}
    \tilde{\epsilon}_\theta = \epsilon_\theta(x_t, \emptyset) + w \cdot \big(\epsilon_\theta(x_t, c) - \epsilon_\theta(x_t, \emptyset)\big).
    \label{eq:cfg}
\end{equation}
More generally, CFG can be viewed as a simple residual guidance rule between a conditional and an unconditional prediction, and this principle is applicable beyond diffusion noise prediction.

Therefore, while TAG adopts a fixed CFG-style residual extrapolation rule, we systematically investigate how to construct a \emph{target-agnostic} visual baseline $I_{\text{uncond}}$ in the visual domain, and evaluate several practical instantiations under different inference settings.

\subsection{TAG Formulation During Inference}
\label{sec:tag_formulation}

TAG contrasts policy predictions under two visual conditions.
Let $I_{\text{cond}}$ denote the original observation that contains full target evidence, manipulation and scene context.
Let $I_{\text{uncond}}$ denote a \emph{target-agnostic} visual baseline that suppresses target-specific cues while retaining (or deliberately controlling) the scene context; in practice $I_{\text{uncond}}$ will be instantiated by several baselines studied in \cref{sec:Abla}.

We denote by $v_\theta(x_t, I)$ the velocity field predicted by our flow-matching policy at state $x_t$ under visual condition $I$.\footnote{The same formulation applies if the policy predicts actions (or action chunks) directly.}
Given the two-branch predictions $v_\theta(x_t, I_{\text{cond}})$ and $v_\theta(x_t, I_{\text{uncond}})$, TAG forms a guided prediction via a CFG-style residual extrapolation:
\begin{equation}
\label{eq:visual_cfg}
v_{\text{TAG}}(x_t)
= v_\theta(x_t, I_{\text{uncond}})
+ w\Big(v_\theta(x_t, I_{\text{cond}}) - v_\theta(x_t, I_{\text{uncond}})\Big),
\end{equation}
where $w$ is the guidance scale.
The residual guidance $\Delta v(x_t)=v_\theta(x_t, I_{\text{cond}})-v_\theta(x_t, I_{\text{uncond}})$ estimates the marginal contribution of target evidence to the policy prediction, and scaling it suppresses clutter-induced nuisance priors while strengthening target-driven action components.

The practical effect of this mechanism is visually confirmed in \cref{fig:vlabench_attentionheatmap}, where applying TAG successfully shifts the model's diffused attention away from distractors and precisely grounds it onto the correct target.

\subsection{Constructing $I_{\text{uncond}}$}
\label{sec:uncond_design}
To implement the visual guidance in Eq.~\eqref{eq:visual_cfg}, we consider three practical instantiations of the target-agnostic baseline $I_{\text{uncond}}$, each capturing different levels of nuisance priors. Specifically, $I_{\text{erase}}$ is employed during training to learn a grounded target-agnostic prior, while $I_{\text{bg}}$ and $I_{\text{black}}$ (referred to as \textbf{TAG-bg} and \textbf{TAG-black}, respectively) serve as the primary baselines during inference. These variants are detailed below:

\paragraph{\textbf{Variant A: background-only image $I_{\text{bg}}$.}}
We construct $I_{\text{bg}}$ by taking a reference frame of the episode and erasing (inpainting) the \emph{entire foreground}, including the manipulated target object, the manipulator, and other movable objects, while keeping the static environment (e.g., table and background) intact.
We keep $I_{\text{bg}}$ fixed for the entire episode to ensure temporal stability.
This variant yields a clean and stable estimate of environment-induced nuisance priors for guidance.

\paragraph{\textbf{Variant B: target-erased image $I_{\text{erase}}$.}}
As an ablation, we also use a target-erased baseline $I_{\text{erase}}$ that removes only the manipulated target object while preserving the manipulator and other scene elements.
This choice is closer to the counterfactual observations used during training (\cref{sec:train_calibration}), but it may retain stronger distractor competition in the visual stream.

\paragraph{\textbf{Variant C: black-only image $I_{\text{black}}$.}}
We further consider a constant black image as $I_{\text{uncond}}$, which intentionally suppresses visual features and provides a simple, temporally stable nuisance baseline.

\paragraph{Deliberate Train--Inference Asymmetry.} 
It is important to emphasize the intentional asymmetry in our design: while training utilizes $I_{\text{erase}}$ to learn a grounded target-agnostic prior under realistic embodiment constraints (see \cref{sec:train_calibration}), inference primarily employs $I_{\text{bg}}$ as $I_{\text{uncond}}$. This asymmetry is strategic: the former stabilizes learning under physically plausible observations, while the latter maximizes robustness by providing a cleaner nuisance estimate that more effectively cancels background- and appearance-induced biases during guidance.

\subsection{Training-time Calibration}
\label{sec:train_calibration}

To ensure the mathematical validity and stability of the two-branch contrast in TAG, we incorporate a training-time calibration procedure inspired by the standard implementation of CFG. In generative modeling, CFG requires jointly training the model on both conditional and null-conditioned inputs to learn an informative guidance direction.

Analogously, we expose the policy to counterfactual, target-agnostic observations during training. Let $I_{\text{erase}}$ denote a target-agnostic observation where the manipulated target object is removed while the manipulator and other scene elements remain. During training, we replace the original observation $I_{\text{cond}}$ with $I_{\text{erase}}$ with a probability $p_{\text{cf}} = 0.1$.

This calibration encourages the policy to produce a stable nuisance prediction when target evidence is absent. Consequently, the residual term in Eq.~\eqref{eq:visual_cfg} can more accurately capture the marginal contribution of the target evidence at inference time, leading to a cleaner and more effective steering signal.

\subsection{Counterfactual Synthesis Pipeline}
\label{sec:counterfactual_synthesis}
This section describes the data construction pipeline used to generate counterfactual observations for TAG.
Concretely, it produces target-erased observations for training calibration (i.e., $I_{\text{erase}}$ in \cref{sec:train_calibration}) and can also be adapted to generate a background-only reference for inference baselines (i.e., $I_{\text{bg}}$ in \cref{sec:uncond_design}), depending on which regions are erased.

To achieve explicit disentanglement between the static environment and manipulation dynamics, we develop an automated pipeline to synthesize counterfactual video sequences where the target object is removed while the background is reconstructed.

\subsubsection{Instruction-Guided Target Semantic Parsing.}
For a given episode, we first sample frames exclusively from the latter half of the video. This temporal heuristic is designed to bypass the initial stages where the robotic arm has not yet engaged with the object, thereby preventing misleading visual context for the model. Subsequently, the sampled frames, along with the corresponding language instruction, are fed into Qwen3-VL~\cite{qwen3vl2025}. Qwen3-VL extracts a precise textual description of the target object, which then serves as the text prompt for the next process to perform accurate bounding box localization.

\subsubsection{Dynamic Masking and Inpainting.}
Next, we employ Grounding DINO~\cite{liu2023grounding} to extract bounding boxes of the target object across the video frames. Serving as spatial prompts, these bounding boxes are input into SAM 2~\cite{ravi2024sam2} to produce continuous spatiotemporal masks throughout the video sequence. We then process these masked regions using the MiniMaxRemover~\cite{zi2025minimaxremovertamingbadnoise} model to perform object erasure and background inpainting.
By choosing the erased regions, the same pipeline can instantiate different counterfactual variants used in TAG (e.g., target-only erasure for $I_{\text{erase}}$, or foreground removal for $I_{\text{bg}}$).
This process synthesizes a target-agnostic sequence that represents the environmental prior, effectively removing the manipulation-centric information while maintaining visual continuity.

\subsubsection{Closed-Loop Quality Assessment.}
To ensure the fidelity of the synthesized priors, we implement a closed-loop automated audit mechanism. The inpainted frames are fed back into the VLM to verify the restoration quality. If the model detects residual target fragments or significant structural artifacts, the sample is rerouted to the Grounding DINO stage for iterative bounding box refinement. This feedback loop allows the pipeline to adaptively correct localization errors that may have caused sub-optimal erasure. We set a maximum threshold of three retry cycles per sample; if a sequence fails to meet the quality criteria after the final iteration, it is flagged as an erasure failure and permanently excluded from the training distribution. This rigorous filtering ensures that the policy is supervised only by high-fidelity, target-agnostic environmental priors.

%% file: sec/4_exp.tex
\section{Experiment}

\begin{figure}[t]
  \centering
  \includegraphics[width=0.9\textwidth]{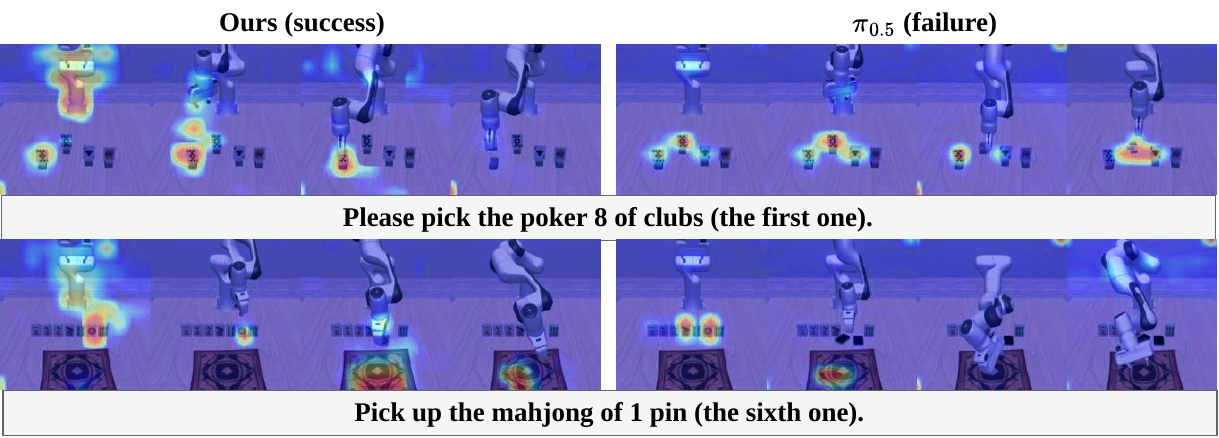}
  \caption{\textbf{Attention map comparison between TAG and $\pi_{0.5}$.} TAG (left) precisely grounds the attention on the target and ignores distractors, resulting in successful manipulation.  $\pi_{0.5}$
  (right) struggles with distractor competition, showing diffused attention that leads to task failure.
  }
  \label{fig:vlabench_attentionheatmap}
  
\vspace{-8pt}
\end{figure}

\subsection{Training Setting}

We adopt $\pi_{0}$~\cite{black2024pi_0} and  $\pi_{0.5}$~\cite{intelligence2025pi05} as our baseline models. Both rely on a Gemma~\cite{team2024gemma} vision-language backbone, with base weights acquired via massive-scale pretraining on diverse robotic datasets. Starting from these rich initial representations, we adapt the architecture to specific tasks through fine-tuning, simultaneously embedding our proposed enhancements.

Each model is trained for a total of $30,000$ steps. The batch size is empirically set to $32$ for the LIBERO and $24$ for VLABench. The optimization process employs a cosine decay learning rate schedule with a warm-up phase of $1,000$ steps. Specifically, the peak learning rate is set to $2.5 \times 10^{-5}$ and gradually decays to a final value of $2.5 \times 10^{-6}$. To ensure training stability and enhance the generalization of the learned weights, an Exponential Moving Average (EMA) with a decay rate of $0.999$ is applied during the training process. Both training and testing are performed using a single NVIDIA RTX PRO 6000 GPU.

\subsection{Simulation Experiment}

We comprehensively evaluate our proposed method across three representative benchmarks. \textbf{LIBERO}~\cite{liu2023libero} provides a standard suite of manipulation tasks specifically designed to evaluate robot learning, and has emerged as one of the most widely adopted benchmarks for evaluating VLA models. To further assess the generalization capabilities and robustness of our model, we incorporate \textbf{LIBERO-Plus}~\cite{fei2025liberoplus}, which introduces enhanced visual and spatial perturbations to the original environments. Finally, we evaluate our model on \textbf{VLABench}~\cite{zhang2024vlabench}, specifically utilizing the representative Track 1, which encompasses a diverse set of complex manipulation scenarios.

\subsubsection{Analysis on LIBERO.}

\input{tables/libero_result}

The empirical results are summarized in \cref{tab-libero_results}. Overall, \textbf{TAG-bg} achieves the highest average success rate (SR), outperforming the baseline model $\pi_{0.5}$ and other powerful models (InternVLA-M1~\cite{chen2025internvla} and GR00T-N1.6~\cite{bjorck2025gr00t}). While the baseline $\pi_{0.5}$ already achieves near-saturation performance on the Spatial, Object, and Goal tasks, our TAG mechanism still manages to consistently push these boundaries higher. 

Most notably, our method yields the most significant improvements on the challenging Libero-Long category, which requires executing complex, long-horizon manipulation sequences. The $\pi_{0.5}$ baseline achieves an $89.6\%$ SR on this split. By applying \textbf{TAG-bg}, we boost this performance to $97.0\%$. This improvement validates our hypothesis: explicitly factoring out background visual distractors enables the model to maintain highly focused attention on the manipulated objects and end-effector actions, which is particularly critical for the compounding errors typically seen in long-horizon tasks.

\input{tables/libero_plus_result}

\subsubsection{Analysis on LIBERO-Plus.}
To evaluate the robustness of our method against out-of-distribution visual and dynamic perturbations, we conduct experiments on the challenging LIBERO-Plus. As detailed in \cref{tab-libero_plus_results}, the benchmark tests model resilience across seven distinct perturbation settings. 

Our proposed \textbf{TAG-bg} and \textbf{TAG-black} achieve the second-highest and the highest average success rates, respectively, among all compared methods. Compared to the $\pi_{0.5}$ baseline, TAG yields significant improvements across nearly all perturbation categories. Most notably, under severe visual domain shifts such as ``Camera'' (viewpoint changes) and ``Noise'' (sensor artifacts), our TAG strategy demonstrates clear robustness improvements.  This compellingly demonstrates that by explicitly teaching the model to differentiate between the active foreground (manipulator and object) and the static background, the policy becomes inherently more robust to environmental variations and visual distractors.

\subsubsection{Analysis on VLABench.}
To further validate our method's efficacy on fine-grained and highly complex manipulation tasks, we analyze performance on VLABench Track 1. This track features visually dense scenes, such as selecting specific items (Poker, Mahjong) from a cluttered workspace. We report both the process score (PS) to evaluate intermediate action precision and the overall SR.

\begin{figure}[t]

  \centering
  \includegraphics[width=0.6\textwidth]{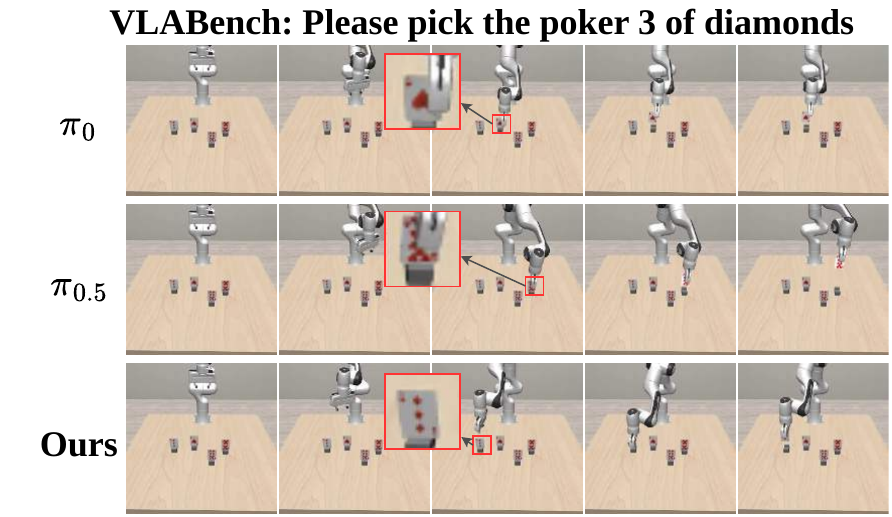}
  \caption{
  \textbf{Comparison on the VLABench benchmark.}
  For tasks involving highly similar distractors (e.g., ``Please pick the poker 3 of diamonds''), baseline policies ($\pi_0$ and $\pi_{0.5}$) are misled by visually similar distractors and grasp the wrong cards. By leveraging the TAG strategy, our method effectively filters out these distractions, shifting attention precisely to the correct target object to prevent misjudgments and ensure an accurate grasp.
  }
  \label{fig:taskshow_vlabench}
  
\vspace{-8pt}
\end{figure}

As shown in \cref{tab-vlabench_sr_results} and \cref{tab-vlabench_ps_results}, TAG significantly elevates the baseline performance. While the raw $\pi_{0.5}$ struggles with visual distractors, TAG increases the SR to $53.06\%$ for ``Select Poker'' and $58.16\%$ for ``Select Mahjong.'' A qualitative comparison is provided in \cref{fig:taskshow_vlabench}. Overall, applying \textbf{TAG-bg} achieves an average SR of $55.41\%$, more than doubling the baseline's $29.40\%$. Furthermore, the average PS increases significantly, indicating that our guidance mechanism not only improves task completion but also corrects the intermediate trajectory generation. The profound improvement confirms that in complex, multi-object scenarios, the ability to suppress the background via TAG is crucial for isolating the target object and executing high-precision manipulation. To further investigate the underlying mechanism, we visualize the internal attention maps during the VLABench evaluation in \cref{fig:vlabench_attentionheatmap}. As illustrated, while the baseline $\pi_{0.5}$ policy exhibits diffused attention across visually similar distractors, our TAG method successfully concentrates its focus precisely on the language-instructed target, directly corroborating the performance gains observed in cluttered environments.

\input{tables/vlabench_sr_result}

\input{tables/vlabench_ps_result}

\subsection{Ablation Study}
\label{sec:Abla}

To thoroughly validate the efficacy of the proposed TAG mechanism and its underlying design choices, we conduct comprehensive ablation studies. We specifically investigate the impact of different visual conditioning strategies and critical hyperparameters on the final generation performance.

\begin{figure}[t]
  \centering
  \includegraphics[width=0.6\textwidth]{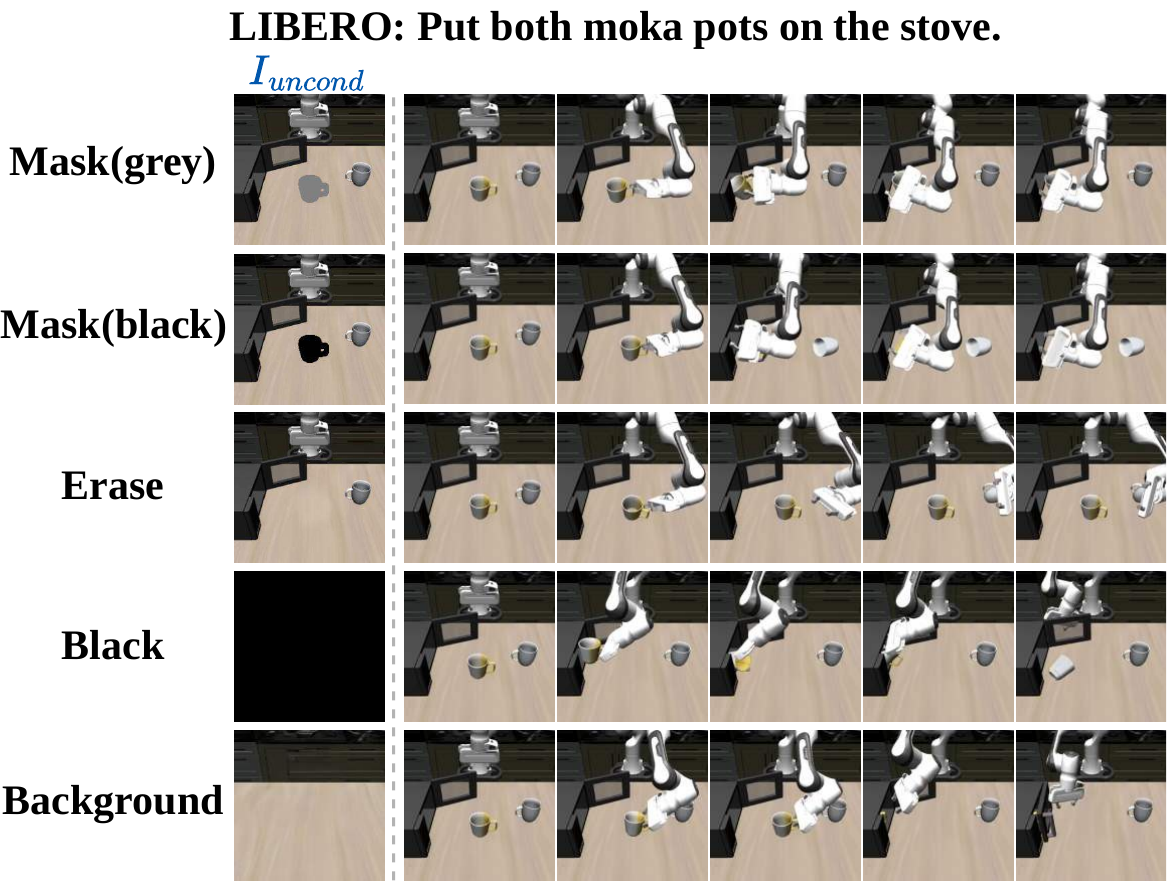}
  \caption{
  \textbf{Visual comparison of different erase methods during inference in Libero.} In the task ``put both moka pots on the stove'', we evaluate various strategies for constructing the unconditional observation to apply TAG. The sequence demonstrates that alternative methods, such as simple masking (Mask-gray/black), inpainting (Erase), and full blackout (Black), disrupt essential spatial priors or introduce visual artifacts, leading to manipulation failures (e.g., colliding with the distractor cup). In contrast, our Background strategy successfully provides a stable and effective unconditional input, enabling the policy to accurately place the moka pot on the stove.
  }
  \label{fig:taskcompare_libero_infer}
  
\vspace{-8pt}
\end{figure}

\input{tables/ablation_on_erase_train}

\subsubsection{Guidance in Training.}
We investigate the impact of different image construction strategies for the unconditional branch during the training phase. As shown in \cref{tab-ablation_erase_train}, training with the synthesized Erase (E) frames yields the most robust performance across various manipulation tasks. Specifically, when evaluated using the Background (BG) method, the model trained with the Erase strategy achieves an optimal average SR of $55.41\%$, significantly outperforming models trained with pure Black (B, $47.01\%$) or static Background (BG, $39.33\%$) images. This indicates that employing explicit object erasure during training forces the model to learn more generalized and resilient representations, effectively preparing it for TAG.

\subsubsection{Erase Method at Inference.}

\input{tables/ablation_on_erase_infer}
We compare five distinct visual condition construction methods in inference: static extraction (using a static background or a pure black image) and three real-time dynamic masking/erasure strategies (real-time-mask (gray), real-time-mask (black), and real-time-erase).
As shown in \cref{tab-ablation_erase_infer}, static extraction methods significantly outperform real-time dynamic strategies. Using a static clean background or a pure black image as the unconditional prompt leads to consistently high SR, whereas frame-wise masking or erasing during inference causes a substantial performance drop. As illustrated in \cref{fig:taskcompare_libero_infer}, dynamic modifications often disturb spatial priors or introduce visual artifacts, which can directly result in manipulation failures such as object collisions.

This performance degradation is particularly pronounced in the Libero-Long task, where the SR collapses from $97.0\%$ to roughly $13\%$. We attribute this severe drop to the temporal inconsistencies and visual artifacts inevitably introduced by real-time dynamic erasure across consecutive video frames. In long-horizon manipulation tasks, these minute, frame-wise visual errors rapidly compound, ultimately destroying the model's coherent understanding of the spatial layout. In contrast, leveraging a static background or a pure black canvas extracted from the initial frame provides an exceptionally stable and clean anchor for the diffusion model. This stable reference allows the model to reliably execute the ``foreground minus background'' residual computation without temporal interference, thereby exhibiting remarkable robustness.

\subsubsection{Guidance Scale.}
Under a fixed number of guidance steps (Step = $10$), we evaluate the impact of different TAG guidance scales. As shown in \cref{tab-ablation_guidance}, there exists a clear peak for the guidance scale. At $w = 1.25$, the model achieves the optimal average performance. This indicates that moderate CFG strategies effectively amplify the model's attention toward the foreground manipulation intent. However, as the guidance scale becomes excessively large, performance gradually regresses; at an extreme value of $w = 15$, the average SR drops to $51.65\%$. We hypothesize that overly aggressive guidance distorts the natural action distribution manifolds learned by the model during training, leading to less smooth and less feasible generated trajectories.

\input{tables/ablation_on_guidance}

%% file: tables/libero_result.tex
\begin{table*}[t]
\centering
\caption{
\textbf{Task success rates in LIBERO task.} Values in parentheses indicate the absolute improvement in success rate (SR) over the corresponding baseline under identical settings. All metrics are reported as percentages (\%), with the best results highlighted in \textbf{bold}. 
}
\label{tab-libero_results}
\resizebox{0.6\textwidth}{!}{
\begin{tabular}{c|cccc|c}
\toprule
\multirow{2}{*}{\textbf{Model}} & \textbf{Spatial} & \textbf{Object} & \textbf{Goal} & \textbf{Long} & \textbf{Average} \\
& SR (\%) $\uparrow$ & SR (\%) $\uparrow$ & SR (\%) $\uparrow$ & SR (\%) $\uparrow$ & SR (\%) $\uparrow$ \\
\midrule
Diffusion Policy~\cite{chi2023diffusion} & 78.3 & 92.5 & 68.3 & 50.5 & 72.4 \\
TraceVLA~\cite{zheng2024tracevla}                     & 84.6 & 85.2 & 75.1 & 54.1 & 74.8 \\
Octo \cite{mees2024octo}                              & 78.9 & 85.7 & 84.6 & 51.1 & 75.1 \\
MDT~\cite{reuss2024multimodal}                        & 78.5 & 87.5 & 73.5 & 64.8 & 76.1 \\
OpenVLA~\cite{kim2024openvla}                         & 84.7 & 88.4 & 79.2 & 53.7 & 76.5 \\
SpatialVLA~\cite{qu2025spatialvla}                    & 88.2 & 89.9 & 78.6 & 55.5 & 78.1 \\
Dita~\cite{hou2025dita}                               & 84.2 & 96.3 & 85.4 & 63.8 & 82.4 \\
CoT-VLA~\cite{zhao2025cot}                            & 87.5 & 91.6 & 87.6 & 69.0 & 83.9 \\
$\pi_0$ FAST \cite{pertsch2025fast}                   & 96.4 & 96.8 & 88.6 & 60.2 & 85.5 \\
GR00T-N1~\cite{bjorck2025gr00t}                       & 94.4 & 97.6 & 93.0 & 90.6 & 93.9 \\
$\pi_0$~\cite{black2024pi_0}                          & 96.8 & 98.8 & 95.8 & 85.2 & 94.2 \\
InternVLA-M1~\cite{chen2025internvla}                 & \textbf{98.0} & 99.0 & 93.8 & 92.6 & 95.9 \\
GR00T-N1.6~\cite{bjorck2025gr00t}                     & 97.7 & 98.5 & \textbf{97.5} & 94.4 & 97.0 \\
$\pi_{0.5}$~\cite{intelligence2025pi05}               & 95.4 & 98.4 & 97.2 & 89.6 & 95.2 \\
\midrule
\rowcolor{gray!20}
$\pi_{0.5}$ + TAG-bg (ours)                            & 97.8 & 99.2 & 97.4 & \textbf{97.0} & \textbf{97.9} \textcolor{red}{(+2.7)} \\
\rowcolor{gray!20}
$\pi_{0.5}$ + TAG-black (ours)                         & 97.6 & \textbf{99.8} & 97.4 & 95.8 & 97.7 \textcolor{red}{(+2.5)} \\
\bottomrule
\end{tabular}}
\vspace{-8pt}
\end{table*}

%% file: tables/libero_plus_result.tex
\begin{table*}[t]
\centering
\caption{
\textbf{Robustness evaluation on the LIBERO-Plus benchmark.} Values in parentheses indicate the absolute improvement in success rate (SR) over the corresponding baseline under identical settings. All metrics are reported as percentages (\%), with the best results highlighted in \textbf{bold}.
}
\label{tab-libero_plus_results}
\resizebox{0.90\textwidth}{!}{
\begin{tabular}{c|ccccccc|c}
\toprule
\multirow{2}{*}{\textbf{Model}} & \textbf{Camera} & \textbf{Robot} & \textbf{Language} & \textbf{Light} & \textbf{Background} & \textbf{Noise} & \textbf{Layout} & \textbf{Average} \\
& SR (\%) $\uparrow$ & SR (\%) $\uparrow$ & SR (\%) $\uparrow$ & SR (\%) $\uparrow$ & SR (\%) $\uparrow$ & SR (\%) $\uparrow$ & SR (\%) $\uparrow$ & SR (\%) $\uparrow$\\
\midrule
OpenVLA~\cite{kim2024openvla}              & 0.8  & 3.5  & 23.0 & 8.1  & 34.8 & 15.2 & 28.5 & 15.6 \\
WorldVLA~\cite{cen2025worldvla}            & 0.1  & 27.9 & 41.6 & 43.7 & 17.1 & 10.9 & 38.0 & 25.0 \\
NORA~\cite{hung2025nora}                   & 2.2  & 37.0 & 65.1 & 45.7 & 58.6 & 12.8 & 62.1 & 39.0 \\
UniVLA~\cite{bu2025univla}                 & 1.8  & 46.2 & 69.6 & 69.0 & 81.0 & 21.2 & 31.9 & 43.9 \\
$\pi_0$~\cite{black2024pi_0}               & 13.8 & 6.0  & 58.8 & 85.0 & 81.4 & 79.0 & 68.9 & 53.6 \\
OpenVLA-OFT (w)~\cite{kim2025fine}         & 10.4 & 38.7 & 70.5 & 76.8 & 93.6 & 49.9 & 69.9 & 55.8 \\
$\pi_0$ FAST \cite{pertsch2025fast}        & 65.1 & 21.6 & 61.0 & 73.2 & 73.2 & 74.4 & 68.8 & 61.6 \\
OpenVLA-OFT (m)~\cite{kim2025fine}         & 55.6 & 21.7 & 81.0 & 92.7 & 91.0 & 78.6 & 68.7 & 67.9 \\
OpenVLA-OFT~\cite{kim2025fine}             & 56.4 & 31.9 & 79.5 & 88.7 & 93.3 & 75.8 & 74.2 & 69.6 \\
RIPT-VLA~\cite{tan2025interactive}         & 55.2 & 31.2 & 77.6 & 88.4 & 91.6 & 73.5 & 74.2 & 68.4 \\
OpenVLA-OFT (plus)~\cite{fei2025liberoplus}& \textbf{92.8} & 30.3 & \textbf{85.8} & 94.9 & 93.9 & \textbf{89.3} & 77.6 & 79.6 \\
$\pi_{0.5}$~\cite{intelligence2025pi05}    & 64.8 & 71.8 & 83.0 & 93.5 & 92.2 & 78.8 & 85.5 & 81.4 \\
\midrule
\rowcolor{gray!20}
$\pi_{0.5}$ + TAG-bg (ours)                & 75.8 & 80.0 & 79.8 & 97.5 & 96.8 & 88.0 & 85.0 & 86.1 \textcolor{red}{(+4.7)} \\
\rowcolor{gray!20}
$\pi_{0.5}$ + TAG-black (ours)             & 77.5 & \textbf{80.5} & 81.5 & \textbf{98.0} & \textbf{97.2} & 89.0 & \textbf{87.0} & \textbf{87.2} \textcolor{red}{(+5.8)} \\
\bottomrule
\end{tabular}}
\vspace{-8pt}
\end{table*}

%% file: tables/vlabench_sr_result.tex
\begin{table*}[t]
\centering
\caption{
\textbf{Success Rate (SR) evaluation on the VLABench benchmark.} We compare the baseline models($\pi_{0}$ and $\pi_{0.5}$) with our proposed TAG variants across five fine-grained selection tasks. Values in parentheses denote absolute SR improvements over the respective baselines. Best results are highlighted in bold.
}
\label{tab-vlabench_sr_results}
\resizebox{0.9\textwidth}{!}{
\begin{tabular}{c|ccccc|c}
\toprule
\multirow{2}{*}{\textbf{Model}} & \textbf{Select Toy} & \textbf{Select Fruit} & \textbf{Select Painting} & \textbf{Select Poker} & \textbf{Select Mahjong} & \textbf{Average} \\
& SR (\%) $\uparrow$ & SR (\%) $\uparrow$ & SR (\%) $\uparrow$ & SR (\%) $\uparrow$ & SR (\%) $\uparrow$ & SR (\%) $\uparrow$ \\
\midrule
\multicolumn{7}{c}{$\pi_0$~\cite{black2024pi_0}} \\
\midrule
base                            & \textbf{54.00} & 48.00 & 16.00 & 6.00  & 6.98  & 26.20 \\
\rowcolor{gray!20} + TAG-bg (ours)     & 48.00 & 50.00 & 36.00 & 48.00 & 15.22 & 39.44 \textcolor{red}{(+13.24)} \\
\rowcolor{gray!20} + TAG-black (ours)  & 52.00 & 50.00 & 40.00 & 34.00 & 11.63 & 37.53 \textcolor{red}{(+11.33)} \\
\midrule
\multicolumn{7}{c}{$\pi_{0.5}$~\cite{intelligence2025pi05}} \\
\midrule
base                            & 28.00 & 44.00 & 26.00 & 32.00 & 17.02 & 29.40 \\
\rowcolor{gray!20} + TAG-bg (ours)     & 38.00 & \textbf{78.00} & \textbf{78.67} & \textbf{53.06} & \textbf{58.16} & \textbf{55.41} \textcolor{red}{(+26.01)} \\
\rowcolor{gray!20} + TAG-black (ours)  & 30.00 & 74.00 & 74.00 & 36.73 & 41.84 & 48.55 \textcolor{red}{(+19.15)} \\
\bottomrule
\end{tabular}}
\vspace{-8pt}
\end{table*}

%% file: tables/vlabench_ps_result.tex
\begin{table*}[htbp]
\centering
\caption{
\textbf{Quantitative results on the VLABench benchmark (Process Score, PS).} We evaluate the performance of baseline models ($\pi_{0}$ and $\pi_{0.5}$) and our proposed TAG variants across five selection tasks. Integrating the TAG strategy consistently yields substantial performance gains for both architectures. Values in parentheses denote absolute PS improvements. Best results are highlighted in \textbf{bold}.
}
\label{tab-vlabench_ps_results}
\resizebox{0.9\textwidth}{!}{
\begin{tabular}{c|ccccc|c}
\toprule
\multirow{2}{*}{\textbf{Model}} & \textbf{Select Toy} & \textbf{Select Fruit} & \textbf{Select Painting} & \textbf{Select Poker} & \textbf{Select Mahjong} & \textbf{Average} \\
&  PS $\uparrow$  &  PS $\uparrow$  &  PS $\uparrow$  &  PS $\uparrow$  &  PS $\uparrow$  &  PS $\uparrow$ \\
\midrule
\multicolumn{7}{c}{$\pi_0$~\cite{black2024pi_0}} \\
\midrule
base                            & \textbf{0.76} & 0.72 & 0.16 & 0.1000 & 0.0814 & 0.3643 \\
\rowcolor{gray!20} + TAG-bg (ours)     & 0.72 & 0.73 & 0.36 & 0.5267 & 0.2283 & 0.5130 \textcolor{red}{(+0.1487)} \\
\rowcolor{gray!20} + TAG-black (ours)  & 0.73 & 0.73 & 0.40 & 0.3933 & 0.2093 & 0.4925 \textcolor{red}{(+0.1282)} \\
\midrule
\multicolumn{7}{c}{$\pi_{0.5}$~\cite{intelligence2025pi05}} \\
\midrule
base                            & 0.60 & 0.67 & 0.26 & 0.3600 & 0.2128 & 0.4206 \\
\rowcolor{gray!20} + TAG-bg (ours)     & 0.60 & \textbf{0.79} & \textbf{0.48} & \textbf{0.7100} & \textbf{0.3800} & \textbf{0.5920} \textcolor{red}{(+0.1714)}\\
\rowcolor{gray!20} + TAG-black (ours)  & 0.54 & 0.75 & \textbf{0.48} & \textbf{0.7100} & 0.3000 & 0.5560 \textcolor{red}{(+0.1354)} \\
\bottomrule
\end{tabular}}
\vspace{-8pt}
\end{table*}

%% file: tables/ablation_on_erase_train.tex
\begin{table*}[t]

\centering
\caption{
\textbf{Ablation study on unconditioning strategies during training and evaluation.} We evaluate the impact of different image processing methods used for TAG: Black (B), Background (BG), and Erase (E). The results indicate that applying the Erase method during training and the Background method during evaluation yields the highest average success rate across all fine-grained selection tasks. Best results are highlighted in \textbf{bold}.
}
\label{tab-ablation_erase_train}

\resizebox{0.9\textwidth}{!}{
\begin{tabular}{cc|ccccc|c}
\toprule
\textbf{Training} & \textbf{Eval} & \textbf{Select Toy} & \textbf{Select Fruit} & \textbf{Select Painting} & \textbf{Select Poker} & \textbf{Select Mahjong} & \textbf{Average} \\
\textbf{ Method} & \textbf{Method} & SR (\%) $\uparrow$ & SR (\%) $\uparrow$ & SR (\%) $\uparrow$ & SR (\%) $\uparrow$ & SR (\%) $\uparrow$ & SR (\%) $\uparrow$ \\

\midrule

B & BG  & 52.00 & 26.00 & 32.00 & 72.00 & 53.06 & 47.01 \\
BG & BG & 50.00 & 38.00 & 20.00 & 62.00 & 26.67 & 39.33 \\
\rowcolor{gray!20}
E & BG & \textbf{60.00} & \textbf{48.00} & \textbf{38.00} & \textbf{78.00} & 53.06 & \textbf{55.41} \\
B & B & 42.00 & 30.00 & 36.00 & 70.00 & \textbf{57.14} & 47.03 \\
BG & B & 52.00 & 38.00 & 34.69 & 68.00 & 34.69 & 45.48 \\
\rowcolor{gray!20}
E & B & 54.00 & \textbf{48.00} & 30.00 & 74.00 & 36.73 & 48.55 \\

\bottomrule
\end{tabular}}
\vspace{-8pt}
\end{table*}

%% file: tables/ablation_on_erase_infer.tex
\begin{table*}[htbp]
\centering
\caption{
\textbf{Ablation study on inference-time unconditional methods for $\pi_{0.5}$.} We compare real-time and non-real-time strategies for generating unconditional inputs. Non-real-time static methods (background and black) significantly outperform real-time masking and erasing, successfully avoiding the severe performance drops observed in long-horizon tasks (Libero-Long). Best results are in \textbf{bold}.
}
\label{tab-ablation_erase_infer}
\resizebox{0.8\textwidth}{!}{
\begin{tabular}{cc|cccc|c}
\toprule
\multirow{2}{*}{\textbf{Real-time}} & \multirow{2}{*}{\textbf{Method}} & \textbf{Spatial} & \textbf{Object} & \textbf{Goal} & \textbf{Long} & \textbf{Average} \\
 & & SR (\%) $\uparrow$ & SR (\%) $\uparrow$ & SR (\%) $\uparrow$ & SR (\%) $\uparrow$ & SR (\%) $\uparrow$ \\
\midrule
\rowcolor{gray!20}
$\times$ & background                & \textbf{97.8} & 99.2 & \textbf{97.4} & \textbf{97.0} & \textbf{97.85} \\
\rowcolor{gray!20}
$\times$ & black                     & 97.6 & \textbf{99.8} & \textbf{97.4} & 95.8 & 97.65 \\
$\checkmark$ & gray-masked           & 87.8 & 76.6 & 77.6 & 12.6 & 63.65 \\
$\checkmark$ & black-masked          & 87.4 & 77.8 & 75.8 & 13.6 & 63.65 \\
$\checkmark$ & erase                 & 87.6 & 79.4 & 77.2 & 13.8 & 64.50 \\
\bottomrule
\end{tabular}}
\end{table*}

%% file: tables/ablation_on_guidance.tex
\begin{table*}[t]

\centering
\caption{
\textbf{Ablation study on TAG guidance scale for $\pi_{0.5}$. }We evaluate the impact of different guidance scales on the Libero benchmark with a fixed 10 denoising steps. Best results are highlighted in \textbf{bold}.
}
\label{tab-ablation_guidance}

\resizebox{0.6\textwidth}{!}{
\begin{tabular}{c|cccc|c}
\toprule
\multirow{2}{*}{\textbf{guidance}} & \textbf{Spatial} & \textbf{Object} & \textbf{Goal} & \textbf{Long} & \textbf{Average} \\
& SR (\%) $\uparrow$ & SR (\%) $\uparrow$ & SR (\%) $\uparrow$ & SR (\%) $\uparrow$ & SR (\%) $\uparrow$ \\
\midrule
    1    & \textbf{98.2} & 98.6 & \textbf{98.6} & 92.8 & 97.05 \\
    1.25 & \textbf{98.2} & 99.2 & 97.8 & 94.6 & \textbf{97.45} \\
    1.3  & 96.4 & \textbf{99.8} & 97.0 & \textbf{95.2} & 97.10 \\
    1.5  & 97.4 & 99.4 & 97.0 & 95.0 & 97.20 \\
    1.8  & 98.0 & 99.6 & 96.0 & 95.0 & 97.15 \\
    2    & 96.2 & 99.0 & 96.8 & 94.0 & 96.50 \\
    3    & 87.8 & 99.0 & 88.4 & 88.6 & 90.95 \\
    8    & 68.2 & 82.6 & 66.0 & 66.6 & 70.85 \\
    15   & 56.0 & 63.0 & 47.4 & 40.2 & 51.65 \\
\bottomrule
\end{tabular}}
\vspace{-8pt}
\end{table*}

%% file: sec/5_conclusion.tex
\section{Conclusion}

In this work, we studied a common failure mode of Vision--Language--Action (VLA) policies in cluttered manipulation scenes: policies often execute feasible actions but bind them to the wrong object instance, resulting in systematic near-miss offsets and wrong-object executions. To address this instance-level grounding problem, we introduced \textbf{TAG} (Target-Agnostic Guidance), a lightweight guidance mechanism inspired by Classifier-Free Guidance (CFG). TAG contrasts policy predictions under the original observation and an object-erased observation to form a residual steering signal that suppresses distractor- and appearance-induced bias while preserving contextual structure for spatial reasoning. Experiments on LIBERO, LIBERO-Plus, and VLABench demonstrate that TAG consistently improves robustness under clutter and reduces instance-level grounding errors without requiring architectural changes to the underlying policy. Future work will explore extending residual guidance to additional controllable factors and improving robustness to imperfect visual observations in real-world settings.


%% file: sec/X_suppl.tex
\clearpage
\setcounter{page}{1}

\setcounter{section}{0}
\setcounter{figure}{0}
\setcounter{table}{0}
\setcounter{equation}{0}

\renewcommand{\thesection}{S\arabic{section}}
\renewcommand{\thefigure}{S\arabic{figure}}
\renewcommand{\thetable}{S\arabic{table}}
\renewcommand{\theequation}{S\arabic{equation}}

\section{Detailed Mathematical Formulation of Target-Agnostic Guidance}
\label{sec:appendix_formulation}

\subsection{Preliminaries: Continuous-Time Flow Matching}
Flow matching formulates generative modeling as the learning of a continuous-time velocity field. Let $\tau \in [0, 1]$ denote the continuous denoising time step, where $\tau=1$ corresponds to the base noise distribution and $\tau=0$ to the target data distribution. Given a clean data sample $x_0$ and a noise sample $\epsilon \sim \mathcal{N}(0, I)$, an intermediate state $x_\tau$ is constructed as:
\begin{equation}
    x_\tau = \tau \cdot \epsilon + (1 - \tau) \cdot x_0
\end{equation}
The model is trained via a regression objective to predict the analytical ground-truth velocity field:
\begin{equation}
    u_\tau = \epsilon - x_0
\end{equation}

\paragraph{\textbf{Inference via ODE Integration.}}
At inference time, data samples are generated by solving the corresponding Ordinary Differential Equation (ODE):
\begin{equation}
    \frac{dx_\tau}{d\tau} = v_\theta(x_\tau, \tau)
\end{equation}
integrating backward from $\tau=1$ to $\tau=0$. This process is deterministic and inherently avoids the need for explicit noise variance schedules.

\subsection{Architecture and Algorithm Pipeline}

\paragraph{\textbf{Training Pipeline.}}
Our objective is to model the conditional distribution $p(A_t \mid o_t)$. Here, $t$ denotes the physical execution time step in the environment, $A_t$ is the continuous action chunk starting from $t$, and $o_t = [I_t^1, \ldots, I_t^n,\, l_t,\, q_t]$ represents the multimodal observation at time $t$. Specifically, $I_t^i$ is the $i$-th RGB image from $n$ camera views, $l_t$ encodes the language instruction, and $q_t$ denotes the robot’s proprioceptive state (e.g., joint angles).

To enable Target-Agnostic Guidance (TAG) during inference, we introduce a counterfactual calibration mechanism. When constructing the observation $o_t$, we replace the original target-present image $I$ with a target-erased image $I_{\text{erase}}$ excluding the wrist camera view with a probability of 10\%.

During training, we randomly sample a denoising time step $\tau \in [0,1]$ and inject standard Gaussian noise $\epsilon \sim \mathcal{N}(0,I)$ into the continuous action:
\begin{equation}
    A_t^{\tau} = \tau\epsilon + (1-\tau)A_t
\end{equation}
Given the noisy action $A_t^\tau$ and the calibrated multimodal observation $o_t$, the network outputs the predicted velocity field $v_\theta(A_t^\tau, o_t)$. The model is optimized by minimizing the Mean Squared Error (MSE) loss to regress the ground-truth velocity:
\begin{equation}
    \mathcal{L}_{\mathrm{MSE}}(\theta) = \mathbb{E}_{\tau, A_t, \epsilon, o_t} \left[ \big\| v_\theta(A_t^\tau, o_t) - (\epsilon - A_t) \big\|^2 \right]
\end{equation}

\paragraph{\textbf{TAG Inference Pipeline.}}
During inference, we initialize the action sequence from pure Gaussian noise, $A_t^{\tau=1}$, and perform multi-step iterative refinement. 

At each integration iteration $i$ of TAG, we construct two distinct observations: a conditional observation $o_t^{\text{cond}}$ incorporating the standard global image $I_{\text{cond}}$, and an unconditional observation $o_t^{\text{uncond}}$ where the global image is replaced by a target-agnostic counterpart $I_{\text{uncond}}$ (e.g., a background-only image $I_{\text{bg}}$) while the wrist image remains unchanged. The model independently evaluates the velocity fields conditioned on these two observations. Subsequently, the guided velocity field is computed via residual extrapolation:
\begin{equation}
\label{eq:visual_cfg}
    v_{\text{TAG}}(A_t^{\tau_i}) = v_\theta(A_t^{\tau_i}, o_t^{\text{uncond}}) + w \cdot \Big( v_\theta(A_t^{\tau_i}, o_t^{\text{cond}}) - v_\theta(A_t^{\tau_i}, o_t^{\text{uncond}}) \Big)
\end{equation}
where $w$ is the guidance scale. This mechanism effectively isolates the action components driven by the target object while suppressing prior biases induced by background distractors.

Finally, a numerical ODE solver utilizes the guided velocity field $v_{\text{TAG}}$ to update the action sequence, integrating from $\tau=1$ toward $\tau=0$. Given a specified number of iterative steps (e.g., $N=10$), the Euler step update is formulated as:
\begin{equation}
    A_t^{\tau_{i+1}} = A_t^{\tau_i} + (\tau_{i+1} - \tau_i) \cdot v_{\text{TAG}}(A_t^{\tau_i})
\end{equation}
After $N$ iterations, the continuous action chunk $A_t$ is fully recovered from the noise and ready for physical execution by the robot.

\section{Detailed Counterfactual Synthesis Pipeline}
\label{appendix:counterfactual_synthesis}

In this section, we elaborate on the implementation details of our automated counterfactual synthesis pipeline. This pipeline is critical for generating both the target-erased video sequences for training calibration ($I_{\text{erase}}$) and the clean background references for unconditional inference ($I_{\text{bg}}$). To ensure scalability and high fidelity, the pipeline incorporates heuristic visual parsing, dynamic boundary constraints, and a closed-loop verification mechanism.

\subsection{Target-Erased Video Synthesis for Training ($I_{\text{erase}}$)}
\label{sec:target-erase}

\begin{figure}[htbp]
  \centering
  \includegraphics[width=0.8\textwidth]{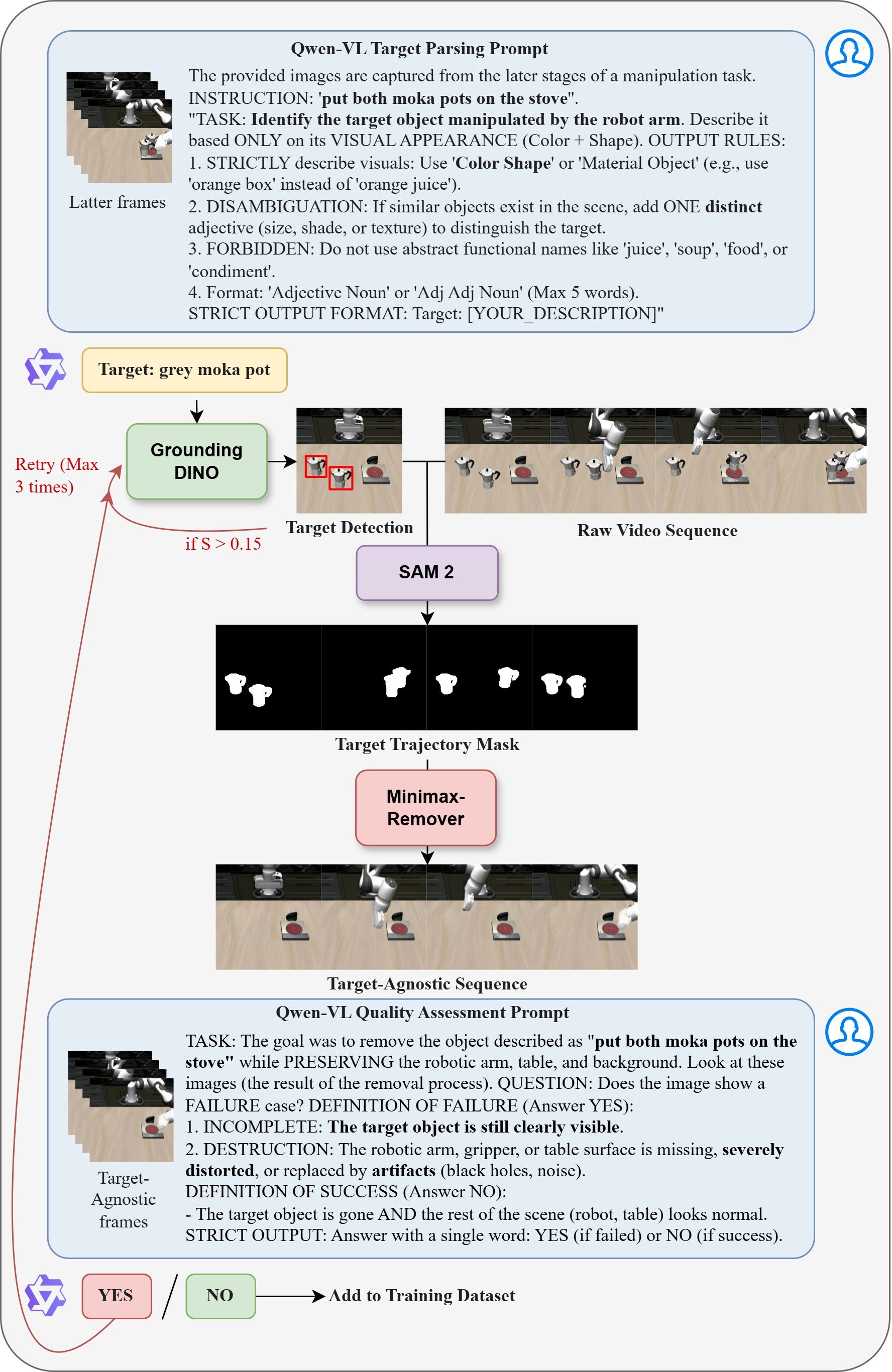}
  \caption{\textbf{Automated Counterfactual Synthesis Pipeline for Target-Erased Video Synthesis.} The framework systematically extracts target-erased videos from raw demonstration videos through a four-stage process: (1) \textbf{Target Parsing}: Qwen3-VL translates abstract task instructions into purely visual descriptions. (2) \textbf{Detection \& Masking}: Grounding DINO localizes the targets. SAM 2 then generates spatio-temporal tracking masks. Here, $S$ denotes the bounding box area relative to the entire frame; (3) \textbf{Inpainting}: Minimax-Remover erases the target object and hallucinates the occluded background. (4) \textbf{Verification}: Qwen3-VL conducts a closed-loop quality assessment to discard sequences with incomplete erasures or destructive artifacts before adding them to the training dataset.
  }
  \label{fig:generate_erase_img}
\end{figure}

As illustrated in \cref{fig:generate_erase_img}, the generation of training data processes the raw video episodes through a robust four-stage pipeline:

\paragraph{\textbf{Heuristic Visual Parsing and Instruction Caching.}} 
Abstract task instructions (e.g., ``clean the table'') and fine-grained objects (e.g., playing cards, mahjong tiles) frequently degrade the recall of open-vocabulary detectors. To address this, we employ heuristic sampling, extracting frames exclusively from the latter half of the video where object interactions are most definitive. We prompt Qwen3-VL~\cite{qwen3vl2025} to extract generic class features. Qwen3-VL is explicitly constrained to ignore fine-grained textures (e.g., numbers on a card) and output purely visual, generalized descriptions (e.g., translating ``poker 3 of diamonds'' to ``red diamond card'').

Furthermore, since identical task instructions within the dataset inherently correspond to the same target objects, forcing Qwen3-VL to evaluate duplicate language instructions repeatedly is computationally redundant. Therefore, we introduce an instruction caching mechanism: the first successfully parsed visual description is serialized and saved into a lightweight JSON key-value store. When the system encounters a recurring task, it directly retrieves the target description from the cache, thereby eliminating redundant VLM inference overhead.

\paragraph{\textbf{Dynamic Detection and Area-Constrained Masking.}} 
The parsed visual prompts are fed into Grounding DINO~\cite{liu2023grounding} for spatial localization. To filter out low-confidence background noise, we implement a dynamic thresholding strategy: the initial confidence threshold is set to 0.35 and is incrementally raised by 0.05 during each subsequent retry loop to progressively tighten the detection criteria.

Crucially, we introduce an \textbf{Area-Constrained Rejection} mechanism. If the total area of the detected bounding boxes exceeds 15\% of the entire frame, it typically indicates a severe misclassification—the detector failed to precisely localize the object and instead erroneously bounded large regions of the background. If such massive masks are permitted into the inpainting phase, the generative model will lack sufficient surrounding context, inevitably filling the region with irregular hallucinations or severe visual artifacts. Consequently, the system immediately intercepts these oversized outputs and triggers a retry.

For valid, high-confidence bounding boxes, SAM 2~\cite{ravi2024sam2} processes them to generate continuous spatio-temporal masks. In multi-target scenarios, each bounding box is registered as an independent tracking instance. To prevent visual artifacts along object boundaries during the subsequent inpainting phase, the aggregated masks undergo morphological dilation using a $5 \times 5$ kernel, ensuring the target boundaries are completely encompassed.

\paragraph{\textbf{Video Spatio-Temporal Inpainting.}} 
The raw video sequence and the dilated spatio-temporal masks are uniformly scaled to a fixed resolution and processed by the MiniMaxRemover model. Over 10 denoising steps, the model synthesizes a continuous, target-agnostic video sequence where the manipulated objects are seamlessly erased, and the occluded background is hallucinated in a physically plausible manner.

\paragraph{\textbf{Quality Verification and Closed-Loop Retry.}}
To prevent poorly inpainted samples from corrupting the training distribution, we deploy Qwen3-VL as an automated quality inspector. Qwen3-VL rigorously evaluates several sampled frames from the inpainted video sequence to verify the success of the erasure process. A synthesis is strictly classified as a failure if: (1) \textbf{Incomplete:} The target object remains partially visible; or (2) \textbf{Destruction:} Critical environmental elements (e.g., the robotic arm, gripper, or table surface) exhibit severe distortion, missing parts, or unphysical artifacts. 
Any failure triggers a retry loop (capped at a maximum of 3 attempts), applying stricter detection thresholds. Episodes that fail all retry attempts are permanently discarded from the training set.

\subsection{Background Reference Synthesis for Inference ($I_{\text{bg}}$)}

\begin{figure}
  \centering
  \includegraphics[width=0.8\textwidth]{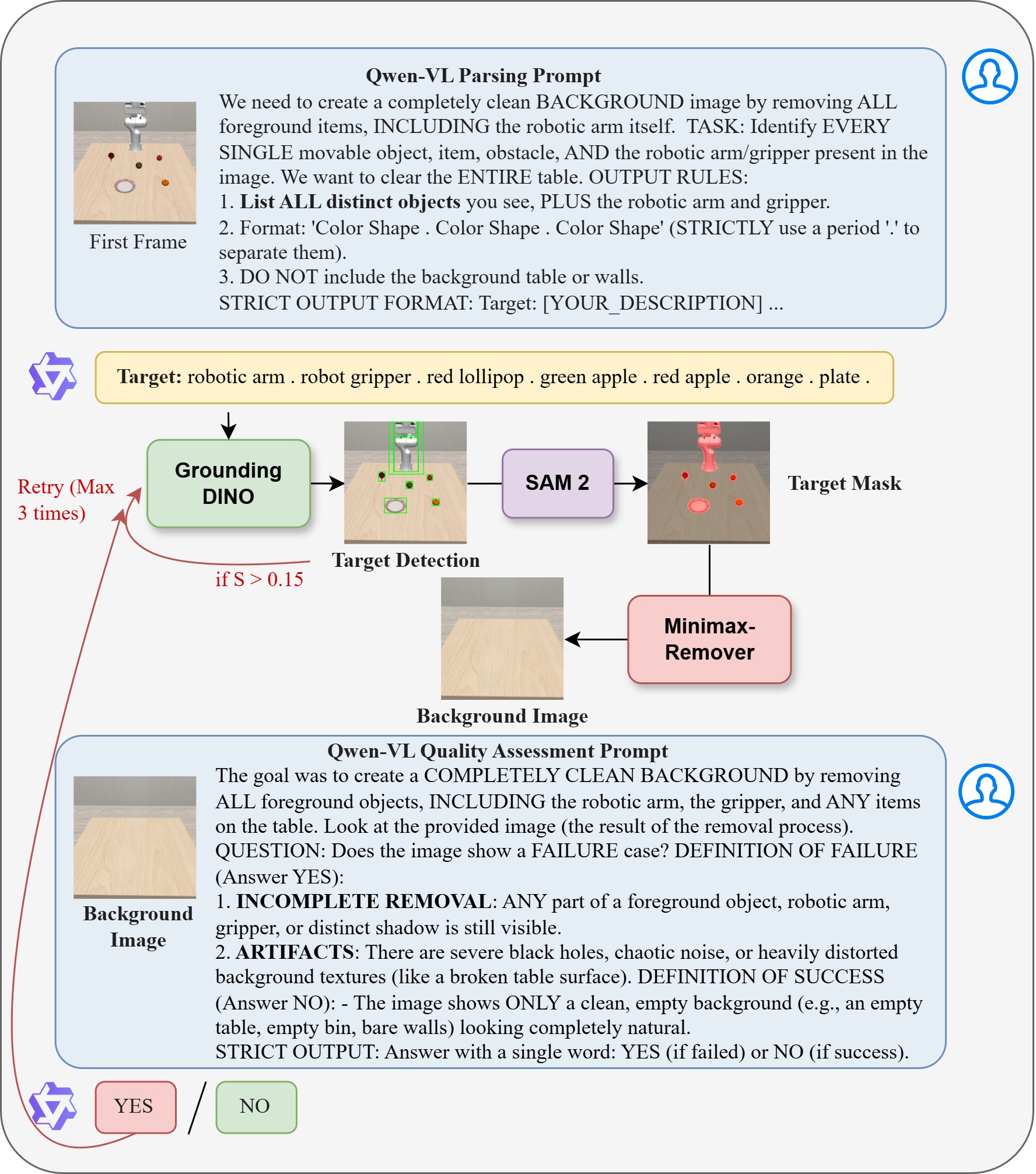}
  \caption{\textbf{Automated Counterfactual Synthesis Pipeline for Background Reference Synthesis.} Adapted from the training pipeline, this process exhaustively parses all foreground objects (including the robot embodiment) from the initial frame ($t=0$). The extracted targets are then processed through the same grounding, inpainting, and verification workflow described above. Here, $S$ denotes the bounding box area relative to the entire frame.
  }
  \label{fig:generate_infer_img}
\end{figure}

Unlike the training phase, the inference baseline requires a completely static, clean background ($I_{\text{bg}}$) containing no movable foreground objects, as illustrated in \cref{fig:generate_infer_img}. To capture the true state of the environment before any physical interaction, we exclusively extract the initial frame ($t=0$) of the episode instead of processing a full video sequence. Operating on this single frame, we adapt our pipeline through the following two modifications:

Since the core synthesis mechanism shares the same paradigm as the training data generation in \cref{sec:target-erase}, we adopt the aforementioned pipeline and only modify the initial Qwen3-VL parsing phase. Instead of extracting task-specific targets, we prompt Qwen3-VL to exhaustively identify every movable object, item, and obstacle in the initial frame, while strictly mandating the inclusion of the robotic arm. This comprehensive list of foreground elements is then fed into the same grounding, area-constrained masking, inpainting, and closed-loop verification pipeline detailed previously to produce the final impeccably clean environmental prior.

\section{Real-World Experiments}
\label{sec:real_world_experiments}
\begin{figure}[t]
  \centering
  \includegraphics[width=0.8\textwidth]{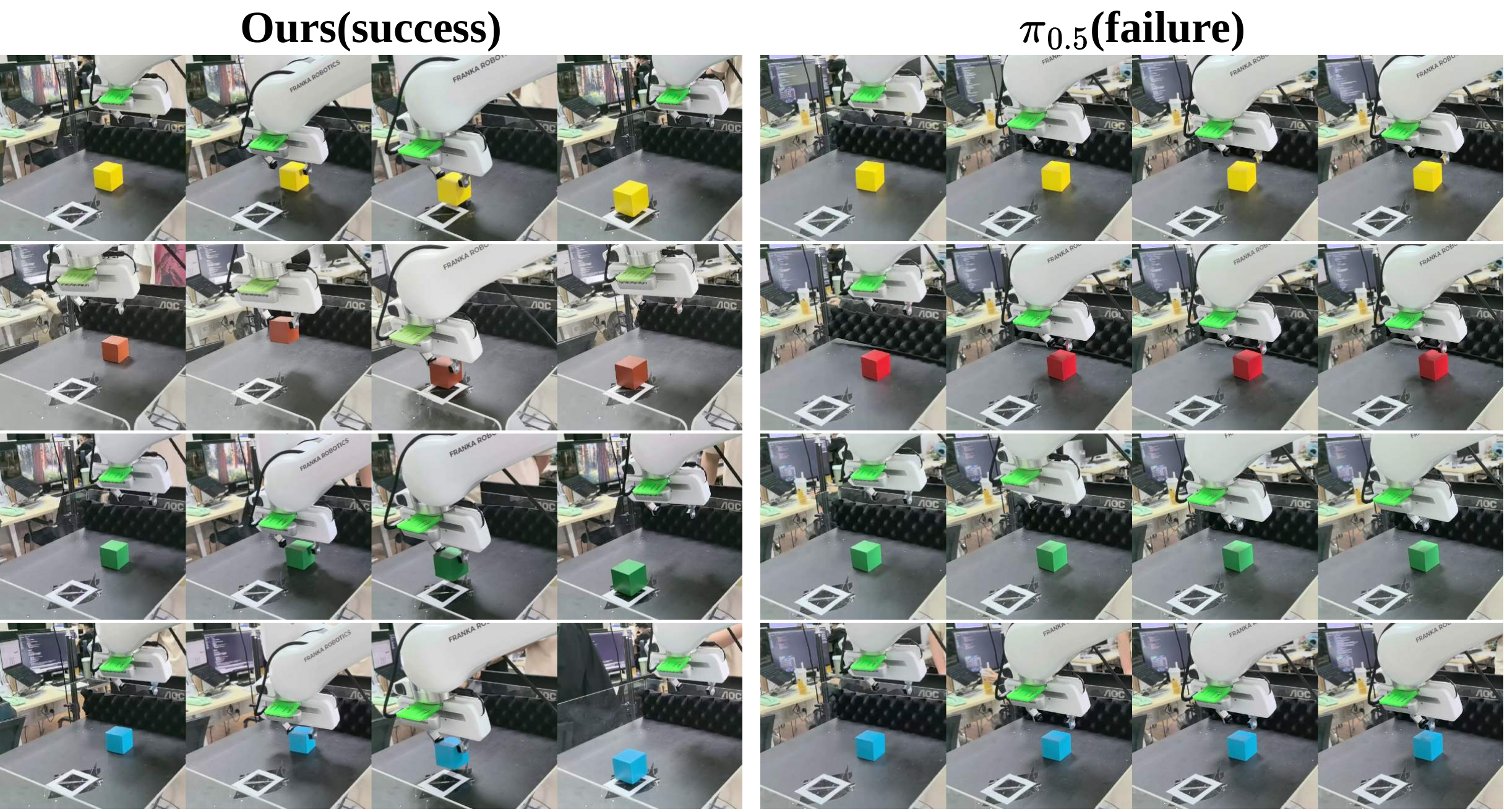}
  \caption{\textbf{Comparison of keyframes in real-world pick-block tasks under camera pose perturbations.} Despite a shift in camera perspective and height, TAG exhibits \textbf{cross-view robustness}, completing the pick-block tasks. In contrast, the baseline $\pi_{0.5}$ is highly \textbf{sensitive to viewpoint variation}, suffering from \textbf{spatial misalignment} that leads to action stagnation above the target.}
  \label{fig:real_compare}
\end{figure}
To validate the spatial generalization and robustness of our proposed method, we conduct manipulation experiments on a real-world Franka Research 3 platform. We utilize the Gello teleoperation framework~\cite{wu2024gello} to collect expert demonstrations.

We design a targeted pick-block task to evaluate the model's visual discrimination and continuous control capability. In this task, the robot must identify a block of a specified color (red, green, blue, or yellow), grasp it, and place it into a designated area. For each color-specific subtask, we collect 50 expert trajectories as the base training data. The corresponding task prompt is defined as: ``Pick up the \{color\} block and place it within the white block area.''

To evaluate robustness under visual distribution shifts, all expert demonstrations are collected under a fixed overhead camera viewpoint, and the policy is trained using this consistent visual configuration. During real-world evaluation, we first conduct experiments under the same base viewpoint, and subsequently introduce random perturbations to the camera pose (e.g., changes in angle and height). Since the models achieve strong performance under the original viewpoint, we deliberately introduce these viewpoint perturbations to create a more challenging evaluation setting. We observe that the baseline model exhibits a notable performance drop under such viewpoint variations.

The qualitative results are presented in \cref{fig:real_compare}. As illustrated, TAG demonstrates superior cross-view robustness. One example result under a perturbed camera viewpoint is shown in the figure. Despite the change in camera perspective, TAG accurately localizes the target and completes the pick task.

In contrast, $\pi_{0.5}$ is highly sensitive to these viewpoint changes. Under the altered testing perspective, $\pi_{0.5}$ frequently suffers from spatial misalignment, leading to action stagnation; the end-effector often hovers near the object but fails to execute the final grasp due to the coordinate distribution shift. This comparison indicates that our approach maintains stable manipulation performance even when the camera viewpoint deviates from the training configuration.